\pdfoutput=1

\documentclass[11pt]{article}

\usepackage[affil-it]{authblk}
\usepackage[]{emnlp2021}
\usepackage{placeins}

\renewcommand{\thefootnote}{\fnsymbol{footnote}}
\usepackage{textcomp}

\usepackage{times}
\usepackage{etoolbox}
\usepackage{hyperref}
\usepackage{amssymb}
\usepackage{mathtools}
\usepackage{multirow}
\usepackage{subcaption, caption}
\usepackage{hyperref}
\usepackage{float}

\usepackage[T1]{fontenc}

\usepackage[utf8]{inputenc}

\usepackage{microtype}

\title{Classification of hierarchical text using geometric deep learning:\\
the case of clinical trials corpus}

\author[ ~1,2]{Sohrab Ferdowsi \thanks{~\,Correspondence: \texttt{sohrab.ferdowsi@hesge.ch} and \texttt{douglas.teodoro@unige.ch } }}
\author[3,4]{Nikolay Borissov}
\author[1]{Julien Knafou}
\author[3,4]{Poorya Amini}
\author[2,1]{Douglas Teodoro}
\affil[1]{HES-SO, Geneva, Switzerland}
\affil[2]{University of Geneva, Switzerland}
\affil[3]{Risklick AG, Bern, Switzerland}
\affil[4]{Clinical Trials Unit, Bern, Switzerland}

\begin{document}

\maketitle

\renewcommand{\thefootnote}{\arabic{footnote}}

\begin{abstract}
We consider the hierarchical representation of documents as graphs and use geometric deep learning to classify them into different categories. While graph neural networks can efficiently handle the variable structure of hierarchical documents using the permutation invariant message passing operations, we show that we can gain extra performance improvements using our proposed selective graph pooling operation that arises from the fact that some parts of the hierarchy are invariable across different documents. We applied our model to classify clinical trial (CT) protocols into completed and terminated categories. We use bag-of-words based, as well as pre-trained transformer-based embeddings to featurize the graph nodes, achieving f1-scores $\simeq 0.85$ on a publicly available large scale CT registry of around 360K protocols. We further demonstrate how the selective pooling can add insights into the CT termination status prediction. We make the source code and dataset splits accessible.
\end{abstract}

\section{Introduction}  \label{sec:intro}

The safety and efficacy evaluation of medications and clinical interventions is performed using clinical trials (CT's)~\cite{plenge2016disciplined}. Prior to their implementation, CT protocols are carefully designed, detailing important aspects of the study, including the number of enrolled patients, their inclusion and exclusion criteria, and the expected outcome, as required by healthcare authorities~\cite{turner2020new}. Regrettably, a large fraction of CT's terminate before reaching a study conclusion~\cite{fogel2018factors}. This is linked directly to delays in providing treatment for the world diseases and to significant excess financial costs~\cite{dimasi2016innovation}. 

\begin{figure}[ht]
\centering
\includegraphics[width=0.56\textwidth]{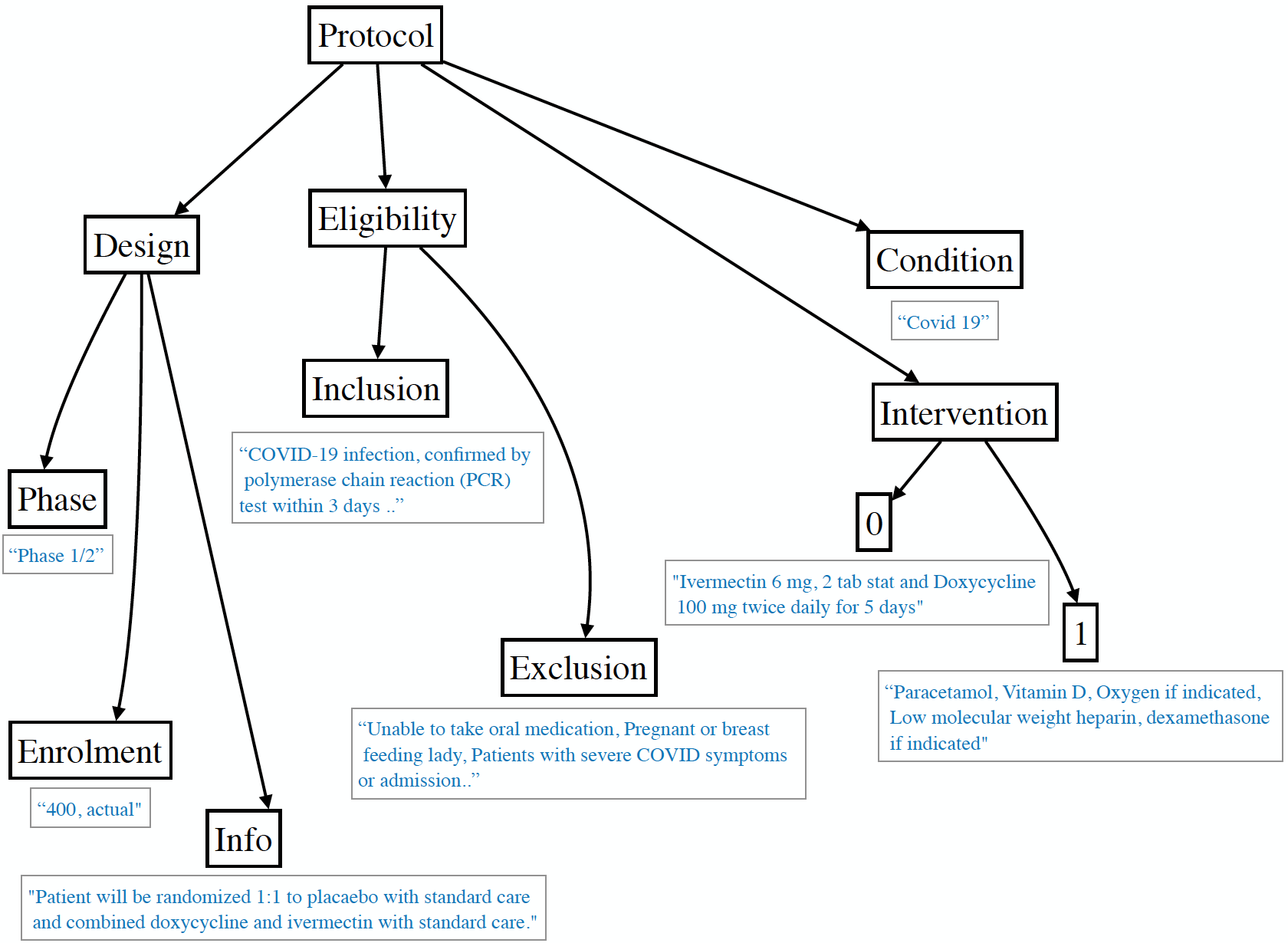}
\caption{Schematic view on the simplified tree structure of a typical clinical trial protocol from ClinicalTrials.gov. Leaf nodes contain free text. Top parent nodes are fixed across trees, but children nodes can contain variable structure.}
\label{fig:graph_example}
\end{figure}

CT protocols are often modelled and represented using tree-like or more generally graph-like structures, such as XML, JSON and DOM~\cite{benson2021uml}. These models use a set of nodes $\mathcal{V}$ representing sections connected by a set of relations $\mathcal{E}$ of type \textit{part of} to encode nested information. The information encoded by a given section of a CT is then the recursive combination of the information encoded by its subsections. As an example, Fig. \ref{fig:graph_example} depicts a simplified CT protocol. Without explicit encoding, a flat-structured text feature extractor would ignore these inter-dependencies. To best consider the inter-connected nature of different elements of a CT protocol, it is thus necessary to take its hierarchical structure into account. 

The systematic way to consider the structural information of data is the promising paradigm of graph neural networks (GNN's) and more generally geometric deep learning \cite{bronstein2017geometric}. Geometric deep learning aims to provide a joint representation of different component features, along with their topology. A key element in most GNN models is the \textit{message passing} algorithm between nodes \cite{gilmer2017neural}, which aggregates the features of nodes based on their neighborhood connectivity. Focusing on the structure of a CT protocol, which is essentially a tree with free-text on its leaf nodes, the parent nodes initialized with zero vectors will aggregate feature vectors of the children nodes during message passing iterations, as depicted in Fig. \ref{fig:steps}.

The power of GNN's lies on their ability to aggregate topology and features when node labeling is arbitrary, i.e., when there is no canonical way of labeling the nodes. Famous examples of this use case are in molecular graphs, e.g., \cite{gilmer2017neural}, when graph nodes consist of atoms, for which no natural ordering is meaningful. The message passing between nodes, followed by a set pooling operation (typically averaging) after node-level representation learning will then provide a global representation for the whole graph, making them suitable for downstream tasks like regression \cite{doi:10.1021/acs.jcim.9b00410}. As for the hierarchical text data like CT protocols, however, this is not exactly the case. While the leaf nodes may have arbitrary structure and hence can benefit from general recipes of GNN's, the parent nodes are typically fixed across all graphs and do not have to undergo global pooling. Therefore, as we will show next, a selective pooling that keeps this structure intact is preferred to the permutation invariant pooling. 

This paper has the following \textit{contributions}: 
\begin{itemize}
    \item This is the first effort to use GNN's on hierarchical text data with free text. While the works of \cite{shang2019gamenet, choi2017gram} use GNN's on one-hot-encoded nodes of medical codes, our node features consist of embeddings of free text, as we discuss in sec. \ref{sec:proposed}. Beyond the example of CT data considered in this paper, this approach can be useful for other hierarchical text data, e.g., like in scientific publications or trademark and patent-related texts.
    
    \item For the hierarchical text data with a combination of fixed and variable node structures, we propose ``selective pooling'' that benefits from the GNN's node-embedding power, as well as the a priori knowledge of the fixed part of the structure.
    \item We present the first deep learning-based approach to CT termination status prediction. The recent work of \cite{elkin2021predictive} uses hand-crafted features and reports AUC $\simeq 0.73$. On a similar experimental setting our results reach AUC $\simeq 0.93$ without feature engineering.     
\end{itemize}
We furthermore provide practical insights and recipes as how to make a bag-of-words (BOW) vectorizer around 50 times faster than a pre-trained BERT model ~\cite{devlin2018bert} in evaluation run time with a loss of F1-score of classification only around $2\%$. 

The rest of the paper is organized as follows. In section \ref{sec:sota}, we describe the related works for the different elements to our paper, i.e., the general problem of text classification, graph-based deep learning, the use of graphs in text, as well as the use of machine learning in the study of CT's. Section \ref{sec:proposed} discusses our proposed solutions in terms of how we featurize the texts and the way we represent hierarchical texts as graphs suitable for graph-based deep learning. Section \ref{sec:experiments} details how we prepare the CT corpus for classification, the baseline and the proposed methods used and a discussion of the classification results of each of these methods, as well as an effort for explainability of the graph-based representation. The paper is finally concluded in section \ref{sec:conclusions}.

\section{Related work}  \label{sec:sota}

\subsection{Text classification} \label{subsec:sota_text_classification}
An extensively studied problem at the core of many NLP applications is the text classification problem, which assigns categorical labels to textual data. Apart from the classification algorithm, feature representation for text is a crucial step of text classification. Classical approaches represent text using the BOW representation of tokens, which disregard the sequential nature of text and essentially provide histogram-like information of tokens within the corpus. A next generation of methods use word embedding techniques, like the word2vec \cite{mikolov2013efficient, mikolov2013distributed} and GloVe \cite{pennington2014glove}, which furthermore consider the neighborhood relation of words and can benefit from external corpora for training the representations. Another line of work focuses on the sequential structure of tokens within text and uses deep learning architectures like the CNN's, as in \cite{kim-2014-convolutional}, or RNN's, as e.g., in \cite{tai2015improved} to capture semantic information.

The state-of-the-art paradigm for a wide range of language understanding tasks, including text classification, is language modeling using transformers \cite{vaswani2017attention}. Most notably BERT~\cite{devlin2018bert} and its subsequent works, like RoBERTa~\cite{liu2019roberta} and XLNet~\cite{yang2019xlnet}, provide state-of-the-art results using different pre-training strategies based on the transformer architecture. These structures benefit from the self-attention mechanism that both provides better sequence modeling capabilities with longer range focus, as well as parallel processing capabilities to fully exploit GPU and TPU processing capacities. The very high expressive power of these networks makes them capable of benefiting from very large corpora trained on different areas and languages providing valuable domain knowledge to the downstream tasks. However, an important shortcoming of the original transformer structures is their inability to process long texts, due to their quadratic complexity w.r.t. the sequence length. Moreover, they require substantial hardware requirements also at the inference time making them not applicable to certain scenarios.

The issue with quadratic complexity of transformers, however, is being actively studied and a multitude of solutions exist to date, such as the Linear transformer of \cite{katharopoulos_et_al_2020}, the Efficient attention \cite{shen2021efficient}, the Linformer \cite{wang2020linformer}, the Longformer \cite{beltagy2020longformer}, or the Reformer \cite{kitaev2019reformer}, among others. While they provide the promise of linear complexity with satisfactory performance, as tested by benchmarks like in \cite{tay2020long}, on the pragmatic side, there still does not exist many pre-trained models available, especially for particular domains like biomedical.

\subsection{Graph neural networks} \label{subsec:sota_graphs}
While most deep learning architectures operate on Euclidean grids with fixed structure, GNN's attempt at the generalization of deep learning concepts to graph structured data using symmetry and invariance.

Consider a graph $\mathcal{G} = (\mathcal{V}, \mathcal{E}; \mathcal{X})$ with node sets $\mathcal{V} = \{ v_1, \cdots, v_{|\mathcal{V}|} \}$, a set of edges $\mathcal{E}$ consisting of pairs of nodes $(u_i, v_i)$, which denote the existence of an edge between the two nodes $u_i, v_i \in \mathcal{V}$, as well as a set of features $\mathcal{X} = \{ \mathbf{x}_1, \cdots, \mathbf{x}_{|\mathcal{V}|} \}$ associated to each of the nodes.

While many applications consider finding useful representations within a graph, the graph classification/regression problems consider finding a global representation $\mathbf{z}_j$ for every given $\mathcal{G}_j \in \{ \mathcal{G}_1, \cdots, \mathcal{G}_N \}$. This should incorporate topology information from $\mathcal{E}_j$, as well as feature information $\mathcal{X}_j$. Note that in general, there does not exist a canonical ordering of nodes within a graph, i.e., within the same graph, nodes can be re-labeled without any semantic implications and no one-to-one correspondence necessarily exists between nodes across different graphs.  

The standard approach to tackle this permutation ambiguity is the message passing between nodes, as e.g., in \cite{gilmer2017neural}, where nodes $v \in \mathcal{N}(u) = \{v \in \mathcal{V} | (v, u) \in \mathcal{E} \}$ in the immediate neighborhood of $u$ send a ``message'' using an ``aggregation'' operation on their features, which is then used to ``update'' the features of $\mathbf{u}$, as:

\begin{equation}
    \mathbf{x}_u^{[l+1]} = \mathbb{U}\left[ \mathbf{x}_u^{[l]} ; \mathbb{A} \left\{ \mathbf{x}_v^{[l]} , \forall v \in \mathcal{N}(u)  \right\} \right],
\end{equation}
where super-scripts $1, \cdots, l, \cdots, L$ refer to the fact that this operation is carried out $L$ times. Starting from $\mathbf{x}^{[1]} = \mathbf{x} \in \mathcal{X}$, $\mathbb{A}\{ \cdots \}$ and $\mathbb{U}[\cdot, \cdot]$ are generic differentiable aggregate and update operations, respectively. Famous examples of these operations are the Graph Convolutional Networks (GCN) from \cite{kipf2016semi}, or the Graph Attention Network (GAT) from \cite{velivckovic2017graph}, among many others.

At the end of $L$ iterations of message passing, each $\mathbf{x}_v^{[L]}, v \in \mathcal{V}$ has aggregated features from its $L$-hop neighbors, so that both topology and feature information have been taken into account. These aggregated features should then follow a global ``pooling'' stage $\mathbb{P}_G\{ \cdots \}$, where a final representation $\mathbf{z}_{\mathcal{G}_j}$ is derived for the whole graph, as:

\begin{equation}
    \mathbf{z}_{\mathcal{G}_j} = \mathbb{P}_G \big\{ \mathbf{x}_v^{[L]}, v \in \mathcal{V}   \big\},
\label{eq:global_pooling}
\end{equation} 
which is usually taken to be simply an averaging operation. This final representation is then treated as an input feature to a generic classifier, usually a differentiable MLP. The learnable parameters of the GNN, i.e., those from the aggregation and update for each layer, as well as the final MLP are then jointly updated with back-propagation using usual techniques of deep learning on mini-batches of training examples $\{ (\mathcal{G}_1, y_1), \cdots, (\mathcal{G}_{|\mathcal{V}|}, y_{|\mathcal{V}|}) \}$, with $y_j$ being the label associated to $\mathcal{G}_j$ for the exemplar case of graph classification.

\subsubsection{GNN's for text}

A new line of work, e.g., \cite{yao2019graph, zhang2020text, ding2020more}, tries to represent textual data as graphs, where the graphical structure is built from co-occurrence of words, either in a corpus level and hence constructing a very big graph for the whole corpus, or in a document level where a separate graph is constructed for each document. Text classification will then be carried out using GNN's as node classification and graph classification problems, for the first and second cases, respectively. This is fundamentally different from our case, where the graph structure is not constructed from text, but the text itself is structured hierarchically, as in a CT protocol.

Another line of work, as in \cite{shang2019gamenet, choi2017gram, shang2019pre}
considers the Electronic Health Records (EHR) data as graphs and uses GNN's to integrate them within healthcare data for solving different tasks. For their case, however, the node features consist of one-hot encoding fixed ontologies, and do not contain free text like in our case.

\subsection{Machine learning efforts on CT understanding} 
\label{subsec: sota_ML_CT}

There has been few works in the literature reporting data-driven methods to assess the termination status of CT's. The work of \cite{follett2019quantifying} uses a simple text mining approach to identify keywords associated to CT termination of the ClinicalTrials.gov (CTGov) data and uses random forests to classify the risk of termination. The work of \cite{geletta2019latent} uses Latent Dirichlet Allocation to find risk-relevant topics and uses the topic probabilities for risk classification using random forests.

The recent work of \cite{elkin2021predictive} poses the problem as a classification of CT's into ``completed'' and ``terminated'' categories. They use feature engineering to feed a set of hand-crafted features into different off-the-shelf classical classifiers. However, even their ensemble methods do not provide satisfactory results. They furthermore perform traditional feature selection and ranking strategies to identify top keywords associated to CT termination.

\section{Proposed method}  \label{sec:proposed}

\subsection{Text featurization}  \label{subsec:featurization}
As a baseline approach to get vectorized representations for free text, we use a BOW-based representation followed by TF-IDF re-weighting, as well as random projections. We also use state-of-the-art pre-trained transformers to improve performance. We next describe these approaches.

\subsubsection{Bag-of-words} 
After standard pre-processing of text (lower-casing, removal of special characters and punctuations, ..), we construct a BOW-based vectorized representation for each protocol, disregarding the hierarchical structures. This is then followed by TF-IDF to re-weight tokens based on their relative importance. 

Concretely, for a set of tokens $\mathcal{W} = \{w_1, \cdots, w_i, \cdots, w_{|\mathcal{W}|}\}$, a CT protocol $1 \leq j \leq N$ is represented by $\mathbf{x}_j = [x_{j1}, \cdots, x_{ji}, \cdots, x_{j|\mathcal{W}|}]^T $, where $x_{ji}$ counts the number of occurrences of $w_i$ in the $j^{\text{th}}$ protocol. TF-IDF re-weights the $i^{\text{th}}$ element of these vectors as 
\begin{equation}
    \tilde{x}_{ji} = \left(\mathbf{x}_j^T\mathbf{1}_{|\mathcal{W}|}\right)  \log{\left(\frac{N}{|| \mathbf{x}(i) ||_0}\right)}  x_{ji},
\end{equation}
where $\mathbf{1}_{|\mathcal{W}|}$ is an all-ones vector of size $|\mathcal{W}|$, the $\ell_0$ norm $||\cdot||_0$ counts the number of non-zero elements of a vector and $\mathbf{x}(i)$ is the $i^{\text{th}}$ row of the matrix $\mathrm{X} = [\mathbf{x}_1, \cdots, \mathbf{x}_j, \cdots, \mathbf{x}_N]$.

An important difficulty with BOW-based representations is the dimensionality $|\mathcal{W}|$, which can be as high as even $10^6$. Feeding this to a model with learnable parameters has a very high chance of over-fitting, as well as a very high computational complexity for matrix-vector operations.

BOW-based representations, however, benefit from very high degrees of sparsity. A classical result from the domain of compressive sensing \cite{candes2005decoding} suggests that a high dimensional sparse vector $\tilde{\mathbf{x}}$ can be projected to lower dimensions using a random matrix $\mathrm{A} \in \Re^{d \times |\mathcal{W}|} $ as $\hat{\mathbf{x}} = \mathrm{A} \tilde{\mathbf{x}}$, virtually without any loss of information. Provided that the sparsity is high enough, one can aim for $ d \ll  |\mathcal{W}|$. Furthermore, it has been shown~\cite{li2006very} that the random projection matrix itself can be chosen to also be sparse. This is very beneficial in practice, since both $\tilde{\mathbf{x}}$ and $\mathrm{A}$ can be stored and multiplied in sparse matrix format, e.g. using numerical packages like SciPy \cite{2020SciPy-NMeth}.

\subsubsection{Pre-trained language models}
A major drawback of the BOW-based representations is that they totally disregard context and the sequential nature of text, since they only provide a histogram-based statistic of token counts. As discussed earlier in sec. \ref{subsec:sota_text_classification}, the state-of-the-art solution to language modeling is based on the transformer architecture \cite{vaswani2017attention}, most notably as in \cite{devlin2018bert}, for which a large number of models pre-trained on very large-scale corpora exist. 

While one gets better performance by further fine-tuning transformers on the downstream task at hand, the computational requirements, most notably their GPU memory consumption, makes the fine-tuning step very expensive for certain tasks. In the case of CT protocols, there is usually more than 100 nodes for each CT, making this step particularly difficult.
We therefore suffice only with fixed embeddings from pre-trained models. 

To embed a piece of text using transformers into a vectorial representation, we use mean-pooling that considers the attention mask for each token into account, as suggested, e.g., in \cite{reimers2019sentence}.
\subsection{Graph representation of hierarchical text}  \label{subsec:proposed_graphs}

\begin{figure*}[ht]
\centering
\includegraphics[width=0.90\textwidth]{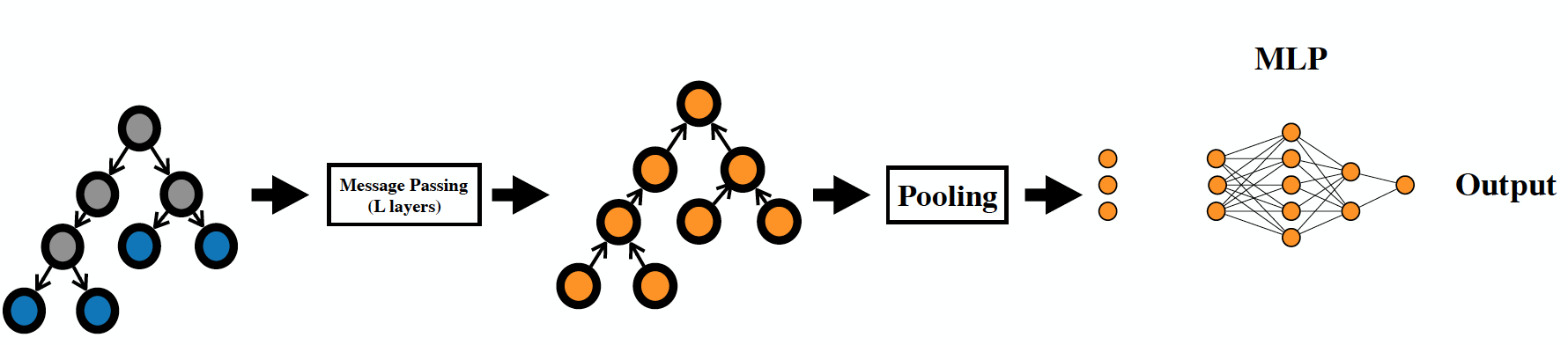}
\caption{A hierarchical structure with extracted feature vectors from free text in leaf nodes is passed to a generic message passing algorithm. The resulting graph with aggregated features is pooled and passed to an MLP to produce the predicted class label.}
\label{fig:steps}
\end{figure*}

Hierarchical text usually comes with a tree structure, where free text appears in the leaf nodes. Compared to the setup of sec. \ref{subsec:sota_graphs} for general graphs, the difference that this brings to the message passing is that the neighborhood $\mathcal{N}(u)$ of node $u$ reduces simply to the set of its children nodes $\mathcal{C}(u)$. Furthermore, the non-leaf nodes $\{u \in \mathcal{V} | \mathcal{C}(u) \neq \emptyset \}$ will be initialized with zero features, and will aggregate features from their children during iterations. Fig. \ref{fig:steps} summarizes the graph representation steps.

\subsubsection{Selective pooling}
Since the general structure of CT protocols is invariable across different CT examples, one can do better than the general pooling strategy of Eq. \ref{eq:global_pooling}.

Consider a set of nodes $\Bar{\mathcal{V}} \in \mathcal{V}$, for which the enumeration is preserved across all $\mathcal{G}_1, \cdots, \mathcal{G}_N$. To benefit from this invariance, one can consider a simple ``selective pooling'' as

\begin{equation}
    \mathbb{P}_S\big[ \mathbf{x}_v^{[L]}, v \in \Bar{\mathcal{V}}   \big] = \underset{v \in \Bar{\mathcal{V}}}{\mathbin\Big\Vert}
     \mathbf{x}_v^{[L]},
      \label{eq:selective_pooling}
\end{equation}
where $\mathbin\Vert$ denotes the concatenation of vectors. To also benefit from the global pooling of GNN's, one can consider the final graph representation as a simple concatenation of global and selective pooling as:
\begin{equation}
    \mathbf{z}_{\mathcal{G}_j} = \mathbb{P}_G\big\{ \mathbf{x}_v^{[L]}, v \in \mathcal{V}   \big\} \mathbin\Big\Vert \mathbb{P}_S\big[ \mathbf{x}_v^{[L]}, v \in \Bar{\mathcal{V}}   \big].
    \label{eq:mixed_pooling}
\end{equation} 
Note that while $\mathbb{P}_G\{ \cdots \}$ is a set operation, $\mathbb{P}_S[ \cdots ]$ is essentially a list operation and hence the order of the elements should be kept uniform across all graphs.



\section{Experiments\footnote{Source code is available at \href{https://github.com/sssohrab/ct-classification-graphs}{https://github.com/sssohrab/ct-classification-graphs}. }}  \label{sec:experiments}

\subsection{Data preparation}  \label{subsec:data}

Healthcare authorities of different countries have different requirements for the publication of the CT protocols. There are 17 CT registries as identified by the WHO\footnote{\href{https://www.who.int/clinical-trials-registry-platform/network}{https://www.who.int/clinical-trials-registry-platform/network}}, where the largest and most complete one is that of \textit{ClinicalTrials.gov} (CTGov). This repository is publicly available to download\footnote{\href{https://ClinicalTrials.gov/AllAPIJSON.zip}{https://ClinicalTrials.gov/AllAPIJSON.zip}} and is the one we use in this paper. The CTGov corpus is updated daily, and the snapshot used in our experiments (downloaded on 10th of December 2020) contains 360,497 CT protocols. Similar to the work of \cite{elkin2021predictive}, we include only interventional studies (i.e., we exclude observational studies as they have a different nature). Furthermore, we exclude the studies with \textit{recruiting} status (as their outcome is not clear yet), and only consider protocols whose overall status is either \textit{completed} (74\% of our subset), which was used to assign the \textit{``completed''} class, or \textit{terminated} (9\%), \textit{withdrawn} (5\%) and \textit{suspended} (1\%), which are grouped collectively into the \textit{``terminated''} class, as these 3 categories have similar risk outcome in practice. This resulting set contains 188,915 protocols, which we split into train, validation and test sets with ratios of $70\%$, $15\%$ and $15\%$, respectively.\footnote{The database, as well as exact splits are available at \textit{zenodo} \cite{ferdowsi_sohrab_2021_5482332}.}

While numerous criteria can be considered to judge a CT as risky (e.g., whether they achieved FDA approval, whether they had reported safety issues, ..), as a basic label-assignment strategy, we consider the ``\textit{completed}'' CT's as low-risk, and the ``\textit{terminated}'' CT's as risky ones. This leaves us with a proportion of $\simeq (74\%, 26\%)$ for low-risk and high risk classes, respectively. 

\subsection{Baseline methods}  \label{subsec:baseline}
We report the classification results based on the following methods.

\textbf{The work of \cite{elkin2021predictive}} described earlier uses a snapshot of the CTGov collection that is downloaded in 2019, to which we do not have access. Furthermore, their labeling strategy slightly differs from ours, as they only consider ``completed'' and ``terminated'' status. Nevertheless, these are very minor differences and the results are still comparable. They report F1-scores $\simeq 0.33$, but only considering the positive class, hence no micro-macro weighting to be included in table \ref{tab:main_results}.

\textbf{Fast-text \cite{joulin2017bag}} is a very efficient library for text classification providing a strong baseline. We used the same pre-processing steps as our BOW-based methods (lower-casing, removal of punctuations and special characters) for tokenization. We trained the model and used the auto-tuning functionality on the validation split using the standard hyper-parameter sets.

\textbf{BOW-W} denotes the standard bag-of-words using a token size of $|\mathcal{W}|$ followed by TF-IDF re-weighting as described earlier in sec. \ref{subsec:featurization}. Note that if the resultant vector representation is fed directly to a classifier with learned parameters, the chance of over-fitting increases with $|\mathcal{W}|$, forcing to chose small vocabulary sizes.

\textbf{BOW-W-RPd} addresses this issue with Random Projections, as described in sec. \ref{subsec:featurization}. As motivated earlier, along with the sparsity of the BOW representations, the projection matrix itself can furthermore be chosen to be highly sparse. In our experiments, we chose $|\mathcal{W}| = 500,000$, $d = 768$ (to be comparable with transformers), and for each row, we set a sparsity of $0.01$ using magnitude thresholding, i.e., only $5,000$ non-zero elements, followed by normalization to unit-norm. This allowed us to significantly speedup the calculations (as well as memory), hence not suffering from the very slow run-time of packages like Gensim (\texttt{models.rpmodel}) \cite{rehurek2011gensim} with our simple SciPy sparse package. As an example, to vectorize 1000 CT protocols, it takes around 7 sec, which is roughly 50 times faster than encoding them with a BERT model (in evaluation mode) on a GPU.

\textbf{PubMedBERT-pretrain-768} refers to the base-BERT pre-trained on PubMed abstracts and PubMedCentral full-texts as introduced in \cite{pubmedbert}. We do not fine-tune the weights on our classification task and use the model in evaluation mode only. We use the interface provided by the Transformer's library \cite{wolf-etal-2020-transformers}, and use the PyTorch framework \cite{NEURIPS2019_9015} in all our experiments. After the embedding vectors are calculated, we use the exact same network as the BOW counterpart (but with different hyper-parameter sets).

\textbf{Flat-1} refers to the case where we take the tree structure of the CT protocol and simply flatten it into 1 field. We then vectorize this field using the above methods and feed it to a classifier network. As for the classifier, we use a 3-layer MLP with a low-rank decomposition of the first linear layer, along with the standard deep learning recipes (batch-normalization, dropout, ..) and use the Adam optimizer with standard hyper-parameters. At each mini-batch, we re-weight the importance of each CT sample to the cross-entropy loss based on the class priors, such that the classes become virtually balanced. We keep this loss function (weighted-BCE) the same across all experiments.

\textbf{Flat-9} summarizes each CT into 9 vectors, each corresponding to one major parent node of the tree. This is to avoid shrinking all information into one vector and keeps some of the original tree structure. The 9 chosen fields are ``sponsor-collaborator'', ``oversight'', ``description'', ``condition'', ``design'', ``arms-intervention'', ``outcomes'', ``eligibility'' and ``contacts-location'' modules of CTGov protocols. When non-existing in some protocols, we assign them all-zero vectors. We feed the resulting 9-channel input tensor to a network with a shared initial small MLP head that independently processes each channel and then concatenates the results and passes it through another small MLP. Except for the concatenation part, this is essentially equivalent to the network used in flat-1.

\textbf{GCN-global} uses 3 layers of the standard graph convolutional block of \cite{kipf2016semi} with hidden dimension of 200 and a global average pooling as in Eq. \ref{eq:global_pooling}. This is then followed by a standard MLP to produce the final output. We use the GCN implementation as provided by the PyTorch-Geometric framework \cite{Fey/Lenssen/2019} and use the data-loading functionalities therein to handle our GNN experiments. 

\textbf{GCN-selective-9} uses the selective pooling that we introduced in Eqs. \ref{eq:selective_pooling} and \ref{eq:mixed_pooling}. To keep the dimensionalities comparable with the global pooling, we chose the output dimension of the third GCN as 20. When the 9 fields, plus the global pooling are concatenated, this will amount to 200, same as in the GCN-global above. In order to see the effect of graph-based modeling, these 9 fields are chosen to be the same as in the Flat-9 method above.

\subsection{Classification results}  \label{subsec:results}
Table \ref{tab:main_results} summarizes the classification results on the test set of our collection based on the standard precision, recall, F1-score macro, F1-score micro, as well as the area under the ROC curve metrics. 

\begin{table*}[!h]
\caption{Performance comparison for models described in sec. \ref{subsec:baseline}. 
}  
\centering 
\begin{tabular}{l|rrrrr}  
\hline                       
Method &Precision & Recall  & \multicolumn{2}{c}{F1-score}  & AUC\\   
& &  & macro & micro & \\
\hline 
\cite{elkin2021predictive} &-  &-  &- & - & 0.7281\\
Fast-text \cite{joulin2017bag} &0.8489 &0.7205 &0.7531 & 0.8402 & 0.8456\\
\hline
BOW-500000-RP768-flat-1 &0.6145 &0.6300 &0.6146 & 0.6453 & 0.7034\\
PubMedBERT-pretrain-768-flat-1 &0.6512 &0.6763 &0.6260 &0.6346 &0.7246 \\
\hline
BOW-1000-flat-9 &0.6489 &0.6713 &0.6488 & 0.6346 & 0.7369\\
BOW-500000-RP768-flat-9 &0.7572 &0.7793 &0.7652 & 0.7906 & 0.8701\\
PubMedBERT-pretrain-768-flat-9 &0.8144 &0.8144 &0.8144 & 0.8419& 0.8911\\
\hline
BOW-500000-RP768-GCN-global &0.8185 &0.8233 &0.8208 & 0.8462& 0.9116 \\
PubMedBERT-pretrain-768-GCN-global &0.8426 &0.8503 &0.8463 & 0.8675 & 0.8881\\
\hline
BOW-500000-RP768-GCN-selective-9 &0.8419 &0.8337 &0.8376 &0.8632 &0.9082 \\
PubMedBERT-pretrain-768-GCN-selective-9 &0.8454 &0.8519 &0.8485 & 0.8697 & 0.9267 \\
\hline
\end{tabular}
\label{tab:main_results}
\end{table*}

The following observations can be made from the classification results:

\begin{itemize}
    \item We notice that taking the hierarchical structure of the CT protocols into account is crucial for classification. The flat-9 models significantly outperform those of flat-1.
    
    \item Increasing the vocabulary size of BOW tokens significantly improves performance. The random projections, as well as the sparsification of the projector matrix are very effective tricks to make this practical.
    
    \item The use of transformer-based embeddings invariably improves performance w.r.t. the BOW. This, however, comes at the price of slower run-times, around 50 times slower than the BOW counterpart starting from raw text to the embedding.
    
    \item GNN-based modeling of CT protocols provides a net increase of performance w.r.t. the flat structures. As a very straightforward example for comparison, because of the linear nature of BOW (disregarding the TF-IDF), the embedded features of each of the fields of the BOW-flat-9 are the summation of their corresponding children nodes of BOW-GCN-selective-9 before starting the message passing. During the algorithm's iterations, the message passing takes the hierarchy into account and pools the information much more effectively than a simple summation, resulting into a superior final performance.
    
    \item The selective pooling proposed in this paper, which incorporates knowledge of the document structure, as well as the message passing of GNN's over the leaf nodes increases performance w.r.t. the global pooling, which benefits only from the latter. This is particularly useful for explainability analyses as we will see next.
 \end{itemize}   
\subsection{Explainability}  \label{subsec:results}  

While attracting a lot of recent attention from deep learning research communities, the explainability of graph neural network models are less explored and hence less developed compared to the grid-structured data like flat text or images \cite{yuan2020explainability}. This is in part due to the lack of locality information which arises from the inherent permutation ambiguity of nodes that we discussed earlier. So the extension of the explainability techniques developed for grid data is not trivial, e.g., due to the non-differentiable nature of the graph adjacency matrix. 

\begin{figure*}[h]
\centering
\includegraphics[width=0.99\textwidth]{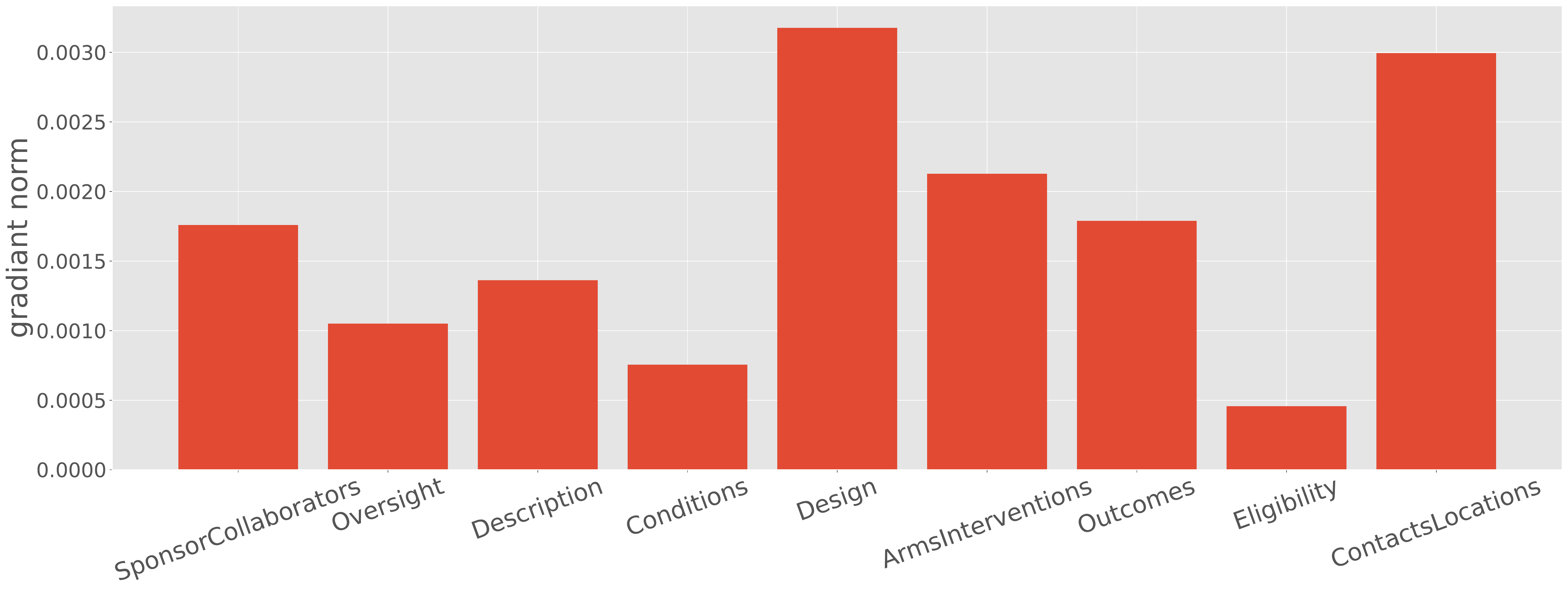}
\caption{Gradient norms of 9 CT protocol fields w.r.t. the class outputs averaged over 1,000 not-completed CT's.}
\label{fig:grad_norms}
\end{figure*}

As a useful workaround to bypass these issues, the selective pooling proposed in this paper can readily use basic gradient-based techniques used in other domains. As an example, we investigate the norm of the gradient of the selectively-pooled nodes w.r.t. the output, i.e.: 

\begin{equation}
    \alpha_v = \Big|\Big|\frac{\partial y^c}{\partial \mathbf{x}_v^{[L]}}\Big|\Big|_2, \;\; v \in \Bar{\mathcal{V}}
    \label{eq:grad_norm}
\end{equation}

Fig. \ref{fig:grad_norms} sketches the average values for 1,000 CT's classified as high-risk by the \textit{BOW-500000-RP768-GCN-selective-9} model above.

\begin{figure*}[!h]
\centering
\subcaptionbox{Per-field\label{subfig:1}} {\includegraphics[width=0.40\textwidth]{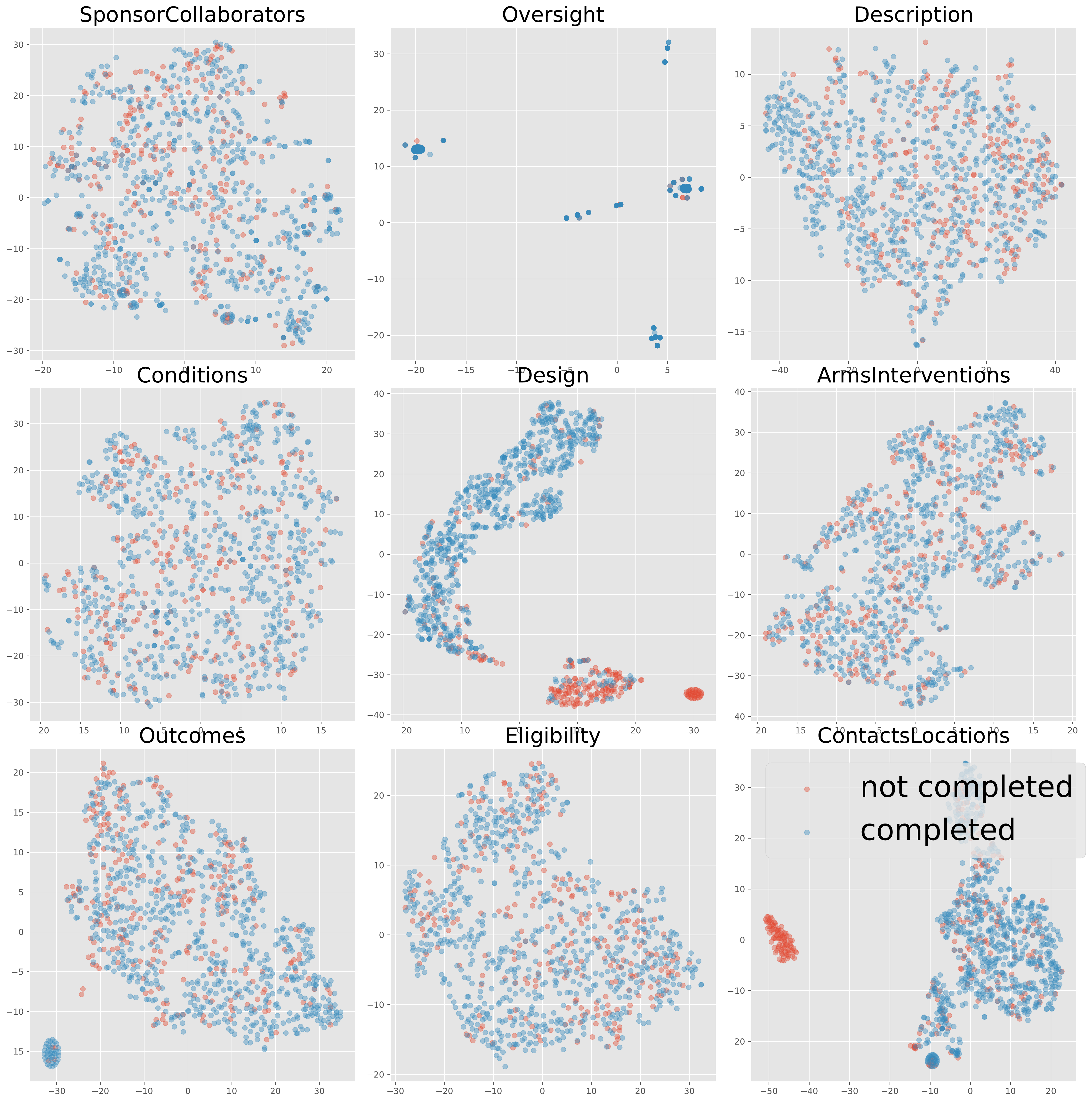}}
   \subcaptionbox{Global\label{subfig:phi}} {\includegraphics[width=0.40\textwidth]{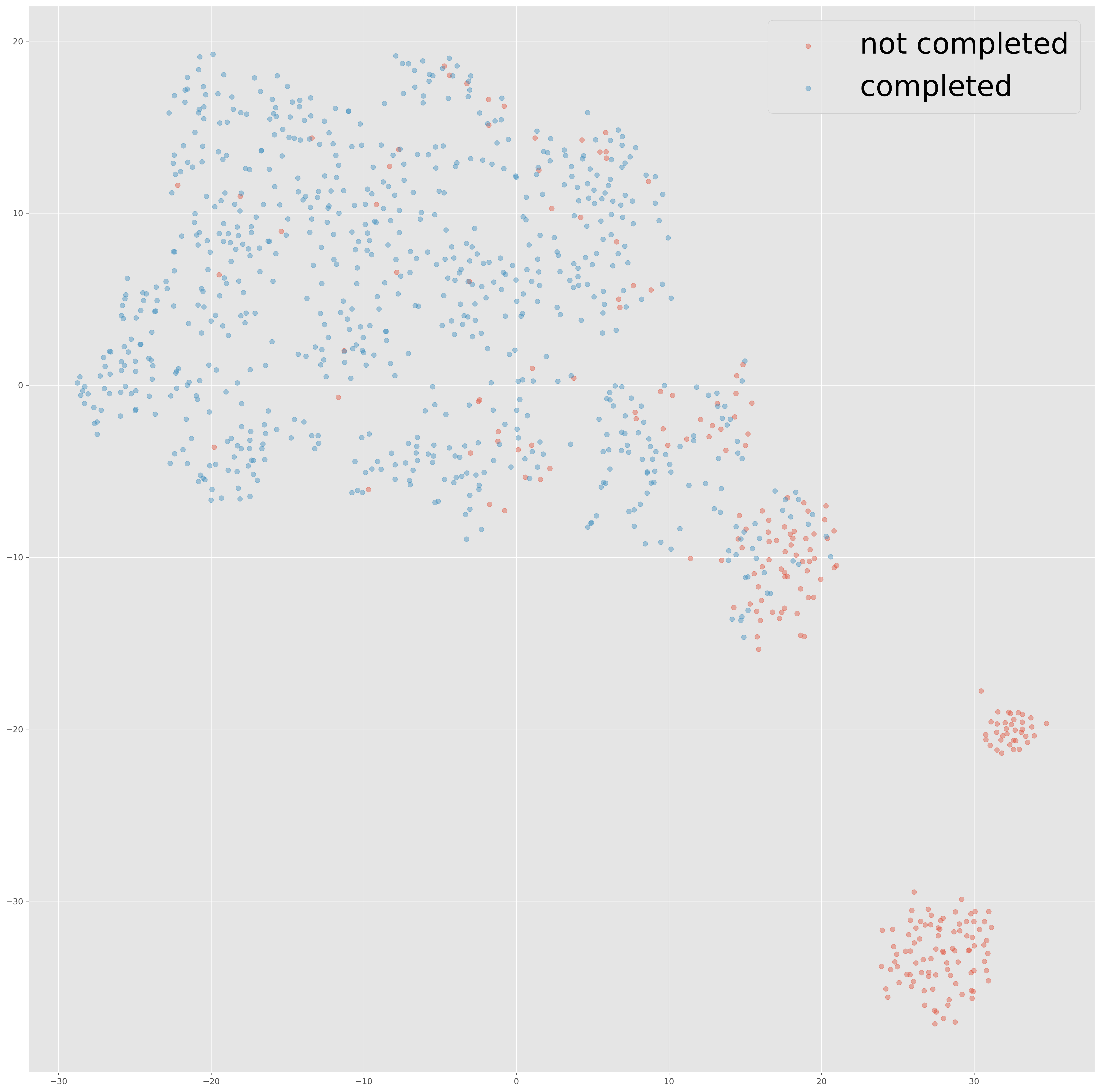}}
\caption{Visualization using t-SNE of the (a) selective pooling of the 9 fields (b) the layer after pooling. The selective pooling reveals the fields usually associated to risk. }
\label{fig:embeddings}
\end{figure*}

Fig. \ref{fig:embeddings} shows the t-SNE \cite{van2008visualizing} visualizations of the same 9 fields ($\mathbf{x}_v^{[L]} \in \Re^{20}$) for the two classes.

The two above figures confirm the source of risk in CT protocols to be the ``Design'' and ``ContactsLocation'' fields. This is in accordance with studies like \cite{fogel2018factors}, which identify the main sources of CT failure as lack of recruitment, that appears under the Design module of the protocol,  and problems related to funding, which can be associated with the locations in which the trials are carried out.


\section{Conclusions}  \label{sec:conclusions}
We used Graph Neural Networks to represent the structure, as well as the extracted features from free text within hierarchical documents. Our use case was the classification of interventional clinical trials into two different risk categories based on their protocols. On a publicly available corpus of around 360K protocols, we showed that the use of GNN's provides a net increase in performance, compared to structure-agnostic baselines. To better incorporate the power of GNN's into the invariable a priori known template, we proposed selective pooling to boost the performance of global pooling. Furthermore, we showed that this approach provides straightforward solutions for explainability, where we demonstrated some consistency between gradient activities of protocol fields within our model to known factors of risk from clinical trials research literature.

\section*{Acknowledgments}
This work is funded by the Swiss Innovation Agency, Innosuisse under the project with funding number 41013.1 IP-ICT.

\FloatBarrier

\bibliographystyle{acl_natbib}
\bibliography{Bibliography}



\end{document}